\documentclass[]{sensenova}

\usepackage{hyperref}
\usepackage{cleveref}
\usepackage{verbatim}

\usepackage{wrapfig}
\usepackage{graphicx}
\usepackage{floatrow}
\usepackage{caption}
\usepackage{subcaption}
\usepackage{listings}
\usepackage{algorithm}
\usepackage{subcaption}

\usepackage{marvosym} 

\usepackage{mmstyles}

\usepackage{tikz}
\usepackage{amssymb}
\usepackage{hyperref}
\usepackage{url}
\usepackage{graphicx} 
\usepackage{wrapfig}
\usepackage{xfrac} 
\usepackage{booktabs, tabularx, colortbl, dashrule, makecell, multirow, caption}  
\usepackage{pifont} 
\newcommand{\cmark}{\ding{51}}
\newcommand{\xmark}{\ding{55}}

\definecolor{cellblue}{RGB}{235,242,248}      
\definecolor{celllavender}{RGB}{241,238,248}  
\definecolor{cellpeach}{RGB}{226,217,243}     

\definecolor{costgreen}{RGB}{235,244,239}   
\definecolor{costblue}{RGB}{235,242,248}    
\definecolor{costorange}{RGB}{249,240,229}  
\definecolor{costrose}{RGB}{248,234,236}    

\usepackage[toc,page,header]{appendix}

\newcommand{\circnum}[1]{%
  \tikz[baseline=(char.base)]{
    \node[shape=circle,draw,inner sep=0.6pt,font=\scriptsize] (char) {#1};
  }%
}

\vspace*{-6mm}
\title{Looped Diffusion Transformer}

\author{
Yong Xien Chng$^{*,\dagger,1,2}$, 
Tianyi Chen$^{*,1,2}$, 
Wenwen Tong$^{1}$, 
Haiwen Diao$^{3}$, 
Zhongang Cai$^{1}$,
\\
Lei Yang$^{1}$, 
Ziwei Liu$^{3}$, 
Lewei Lu$^{1}$, 
Dahua Lin$^{1}$, 
Gao Huang\textsuperscript{\Letter,2}
\\[4px] 
\parbox{\textwidth}{\centering\normalsize
    $*$ Equal Contribution\,
    $\dagger$ Project Lead\,\,
    $\textrm{\Letter}$ Corresponding Author \\[4px] 
    $^1$SenseTime Research \,
    $^2$LeapLab, Tsinghua University \,
    $^3$Nanyang Tehnological Univesity \,
}}

\abstract{
Improving text-to-image models has traditionally relied on increasing model size or the number of denoising steps. In this work, we explore an alternative way to scale computation by repeatedly running shared Transformer blocks within each denoising step, effectively increasing computational depth while keeping the parameter count fixed. This looped computation enables iterative refinement of internal representations without explicit reasoning tokens. However, naive looping fails to consistently improve image quality. We trace this problem to weak supervision across intermediate loops and unregulated attention updates that progressively erode local information. To overcome these challenges, we propose Looped Diffusion Transformer (Looped-DiT), which combines deep supervision across intermediate loops with self-modulating attention to stabilize looped feature updates. Under matched-parameter and matched-compute settings, Looped-DiT consistently outperforms non-looped baselines. Notably, a 260M-parameter looped model can surpass a model 6.5× larger across multiple text-to-image benchmarks while requiring 4.9× lower inference compute. Beyond this performance gain, we find that looped computation can offer a more effective form of iterative computation for diffusion models, with increasing loop depth yielding larger gains than adding more denoising steps under a fixed inference budget. Furthermore, deeper loops can progressively correct mistakes made in earlier loops, exhibiting behaviors suggestive of latent reasoning. Together, these results show that looped computation offers a promising way to scale visual generation models.
}

\date{\today}
\checkdata[Codebase]{\url{https://github.com/OpenSenseNova/Looped-DiT}}

\begin{document}
\maketitle

\begin{figure}[!hb]
    \centering
    \includegraphics[width=\linewidth]{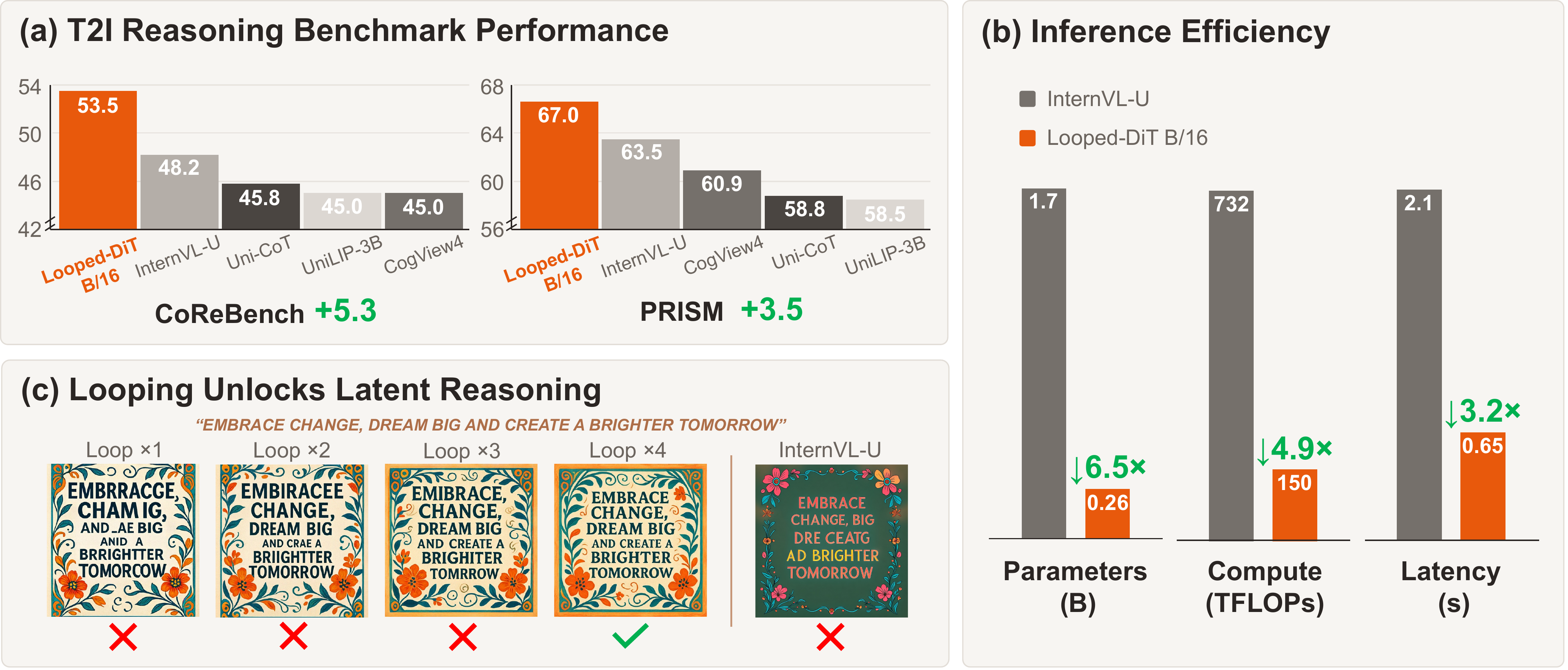}
    \caption{\textbf{Looping is a parameter-efficient way to scale text-to-image generation models.}
    (a) Increasing loop depth greatly improves text-to-image reasoning performance.
    (b) These gains are achieved with far fewer parameters and lower inference cost.
    (c) Self-correction emerges across loops.
    }
    \label{fig:teaser}
\end{figure}


\begin{figure}[t]
    \centering
    \includegraphics[width=\linewidth]{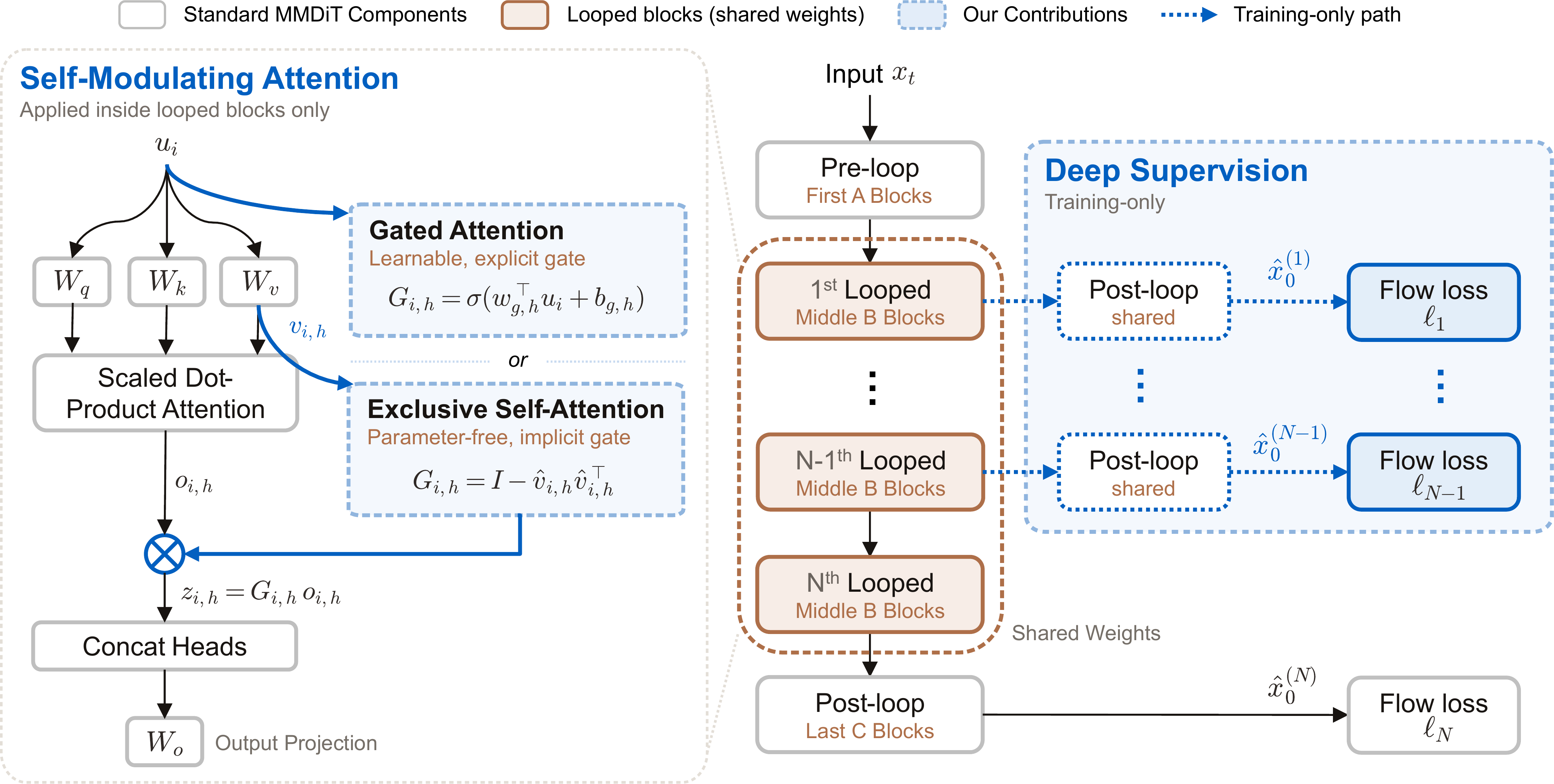}
    \caption{
    \textbf{Overview of Looped-DiT.}
    Shared middle blocks are repeated $N$ times within each denoising step.
    \textit{Self-Modulating Attention} (left) regulates attention updates with token-dependent, headwise gates.
    \textit{Deep Supervision} (right) decodes each intermediate loop output through the shared post-loop blocks, supervising all predictions against the same clean-image target.
    }
    \label{fig:method}
\end{figure}

\section{Introduction}
\label{sec:intro}

Scaling laws~\citep{kaplan2020scaling} show that increasing model size, data, and training compute systematically improves language modeling performance, motivating the search for more efficient ways to scale computation. The two dominant approaches each come with a cost. Deeper or wider Transformers~\citep{vaswani2017attention} require proportionally more parameters, while chain-of-thought reasoning~\citep{wei2022chain} increases inference-time computation through longer token sequences. Looped Transformers~\citep{dehghani2018universal,geiping2026scaling} offer an alternative scaling strategy by repeatedly applying shared middle blocks to refine hidden states, thereby increasing computational depth without increasing parameter count or sequence length. This property is especially attractive for text-to-image generation, where models such as Qwen-Image~\citep{wu2025qwen} and FLUX.2~\citep{flux-2-2025} have grown to billions of parameters, imposing substantial deployment costs. 

The appeal of looping for image generation extends beyond computational efficiency. Generating a coherent image requires more than literal prompt following. The model must \circnum{1} infer the implied visual content, \circnum{2} resolve interdependent constraints, and \circnum{3} identify and correct inconsistencies as generation evolves. Because these processes are inherently visual, looping offers a natural way to perform them directly in hidden representations without explicit textual reasoning. This refinement also integrates naturally with standard diffusion training, since intermediate-loop predictions can be supervised against the same target image without external annotations.

To investigate the potential of looping for text-to-image generation, we introduce it into MiniT2I~\citep{minit2i2026}, a minimal pixel-space Multimodal Diffusion Transformer (MMDiT) with a simple architecture and training pipeline. As shown in Fig.~\ref{fig:method}, we divide its Transformer blocks into three sequential groups, with pre-loop and post-loop blocks surrounding a middle group of looped blocks. Within each denoising step, the pre-loop and post-loop blocks each run once, while the looped blocks run $N$ times with parameters shared across loops to repeatedly update the hidden states. This simple setup allows us to systematically examine whether looping improves performance, characterize the properties that emerge as loop depth increases, and compare looping with alternative strategies for scaling computation.

\begin{figure}[t]
    \centering
    \includegraphics[width=\linewidth]{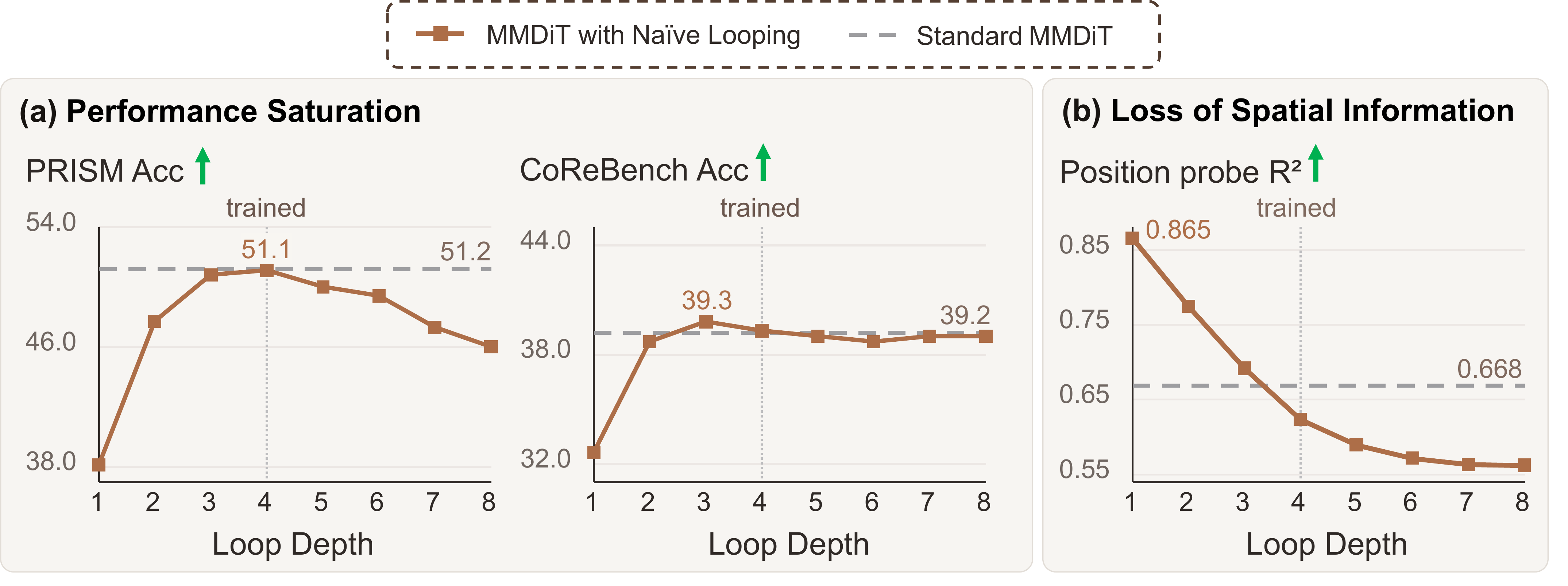}
    \caption{
    \textbf{Limitations of naïve looping in MMDiT.} (a) Increasing inference loop depth beyond that used in training initially improves performance, then saturates and degrades. (b) This degradation coincides with a steady loss of linearly decodable spatial information under a ridge-regression probe.
    }
    \label{fig:naive_looping}
\end{figure}

However, our initial experiments in Fig.~\ref{fig:naive_looping}(a) show that naive looping does not reliably improve generation quality. Performance can remain below the non-looped baseline at shallow loop depths, saturate as more loops are added, and eventually decline beyond the training loop depth. One possible explanation for the performance decline is that repeatedly applying the same transformation produces redundant updates that overwrite or attenuate information in the image-token representations. To probe this hypothesis, we examine how well spatial information is preserved across loops by fitting a ridge-regression~\citep{hastie01statisticallearning} probe at each loop depth to predict each image token's 2D patch-grid coordinates from its hidden state. As shown in Fig.~\ref{fig:naive_looping}(b), $R^2$ drops from 0.865 after the $1^\text{st}$ loop to 0.562 after $8^\text{th}$ loops, an absolute decrease of 0.303, indicating that spatial position becomes progressively less linearly decodable. Together, these findings suggest that \textit{effective looping requires both meaningful supervision of intermediate loop predictions and the ability to adaptively regulate the strength of attention updates as representations evolve across loops}.

To meet these requirements, we introduce Looped Diffusion Transformer (Looped-DiT), combines intermediate-loop Deep Supervision with Self-Modulating Attention to regulate attention updates as representations evolve across loops. To our knowledge, this is the \textit{first} systematic study of looped computation for text-to-image generation. We investigate whether looped computation can improve parameter efficiency, use inference-time compute more effectively, and support latent visual reasoning. Our contributions are summarized as follows:
\begin{enumerate}
\item We demonstrate the effectiveness of looping as a parameter-efficient scaling approach for text-to-image generation. Notably, a 260M-parameter Looped-DiT outperforms non-looped models with roughly $6.5\times$ more parameters across multiple text-to-image benchmarks while requiring $4.9\times$ lower inference compute.
\item We provide evidence for loop depth as a complementary axis for inference-time scaling. Under matched inference compute, allocating additional computation to increasing loop depth yields greater gains than allocating it to more denoising steps.
\item We provide evidence that looping can support latent visual reasoning through iterative hidden-state refinement, with deeper loops progressively correcting earlier errors and resolving interdependent constraints without explicit textual reasoning traces.
\end{enumerate}
\section{Looped Multimodal Diffusion Transformer}
\label{sec:method}


In this section, we describe Looped Diffusion Transformer (Looped-DiT). Looped-DiT is built on MiniT2I~\citep{minit2i2026}, a pixel-space denoiser based on MMDiT~\citep{esser2024scaling}. We use MiniT2I as our backbone because its simple training pipeline provides a controlled setting for studying loop depth. Looped-DiT increases computational depth by repeatedly applying a shared group of Transformer blocks. As shown in Fig.~\ref{fig:method}, we partition the network into a pre-loop stage $\mathcal{A}$, a looped stage $\mathcal{B}$, and a post-loop stage $\mathcal{C}$. Given a noisy image $x_t$ and text condition $y$, let $h_{\mathrm{input}}$ denote the corresponding image and text token representations. The forward computation is
\begin{equation}
h^{(0)} = \mathcal{A}(h_{\mathrm{input}})
\qquad
h^{(r)} = \mathcal{B}(h^{(r-1)}), \quad r=1,\ldots,N,
\qquad
\hat{x}_0 = \mathcal{C}(h^{(N)}).
\label{eq}
\end{equation}
Here, $N$ denotes loop depth. At each iteration, $\mathcal{B}$ updates both image and text hidden states, which become the input to the next iteration. Since the parameters of $\mathcal{B}$ are shared across iterations, increasing $N$ increases effective computational depth without increasing parameter count. This looped computation occurs within each denoising step before the sampler advances, allowing loop depth to be varied independently of the number of denoising steps. To address the limitations of naïve looping, Looped-DiT further incorporates Deep Supervision for intermediate loop predictions and Self-Modulating Attention to regulate attention updates across loops.

\subsection{Deep Supervision}
\label{sec:deep_supervision}

Supervising only the final prediction requires gradients to backpropagate through all subsequent iterations to reach earlier loop states. As loop depth increases, this creates a long, indirect optimization path that deprives earlier states of direct learning signals. Inspired by prior work~\citep{lee2015deeply}, we introduce Deep Supervision, which applies the flow-matching objective to predictions at every loop depth rather than relying solely on the final state.

Specifically, for each loop depth \(n=1,\ldots,N\), we decode the corresponding hidden state through the shared post-loop stage as
\[
\hat{x}_0^{(n)} = \mathcal{C}\!\left(h^{(n)}\right).
\]
We express the flow-matching loss~\citep{lipman2022flow} in terms of clean-image prediction. Given a training image \(x_0\), we construct the noisy input as
\[
x_t = t x_0 + (1-t)\tilde{\epsilon},
\qquad
\tilde{\epsilon} \sim \mathcal{N}\!\left(0,\sigma_{\mathrm{noise}}^2 I\right),
\qquad
t \in (0,1).
\]
All loop predictions share the same noisy input and timestep, and each is supervised against the same clean image $x_0$ using
\begin{equation}
\ell_n = \mathbb{E}\left[
    \frac{\|\hat{x}_0^{(n)} - x_0\|_2^2}{d_x\,c(t)^2}
\right]
\quad
c(t) = \max\{1-t,\tau\}.
\label{eq:loop_loss}
\end{equation}
Here, $d_x$ is the number of image elements, and $\tau>0$ prevents the loss weight from diverging as $t$ approaches one. The overall training objective combines supervision across loop depths as
\begin{equation}
    \mathcal{L} = \sum_{n=1}^{N} w_n \ell_n,
    \quad w_n \geq 0,
\label{eq:loop_supervision}
\end{equation}
where $w_n$ controls the supervision strength at loop $n$. Because intermediate predictions reuse the shared post-loop stage $\mathcal{C}$ and are decoded only during training, Deep Supervision introduces no additional parameters or inference overhead.

\subsection{Self-Modulating Attention}
\label{sec:attention_modulation}

Standard softmax attention controls the relative contributions of source tokens but not the strength of the resulting update. As the same shared blocks are repeatedly applied across loops, their attention updates can become redundant, potentially overwriting or attenuating image-token representations. We therefore use Self-Modulating Attention to regulate repeated updates across loops based on the current hidden states. We study two realizations of this idea in this work: Gated Attention~\citep{qiu2026gated}, which explicitly scales each attention-head output with a learned gate, and Exclusive Self Attention~\citep{zhai2026exclusive}, which applies a state-dependent projection that removes the component along the token's own value direction. We briefly describe both mechanisms below and provide further details and analysis in Appendix~\ref{app:sma}.

For a token $i$ in either modality, let $u_i$ denote its normalized hidden state. The output of attention head $h$ is \(o_{i,h}=\sum_j \alpha_{ij,h}v_{j,h}\), where $\alpha_{ij,h}$ is the attention weight assigned to source token $j$ and $v_{j,h}$ is its value vector. The sum runs over both image and text tokens. Both mechanisms modulate the resulting head output as \(z_{i,h}=G_{i,h}o_{i,h}\) before head concatenation and output projection, where $G_{i,h}$ denotes the corresponding modulation factor.

For Gated Attention, the modulation $G_{i,h}$ is a token-dependent scalar gate,
\begin{equation}
    G_{i,h}^{\mathrm{gate}}
    =
    \sigma\left(
        w_{g,h}^{\top} u_i + b_{g,h}
    \right),
    \label{eq:gated_attention}
\end{equation}
where $w_{g,h}$ and $b_{g,h}$ are modality-specific gate parameters. The resulting scalar explicitly controls the strength of each head contribution to the residual update.

For Exclusive Self Attention, the modulation $G_{i,h}$ instead takes the form of a projection operator that removes the component along the token's own value direction,
\begin{equation}
G_{i,h}^{\mathrm{xsa}}
=
I - \hat{v}_{i,h}\hat{v}_{i,h}^{\top}
\qquad
\hat{v}_{i,h}
=
\frac{v_{i,h}}{\lVert v_{i,h} \rVert_2}.
\label{eq:exclusive_self_attention}
\end{equation}
Applying this projection to the attention output eliminates the self-value term, yielding
\begin{equation}
    z_{i,h}^{\mathrm{xsa}}
    =
    G_{i,h}^{\mathrm{xsa}}
    \sum_{j\neq i}
    \alpha_{ij,h} v_{j,h}.
\end{equation}
Unlike Gated Attention, which explicitly controls update magnitude through a learnable gate, XSA regulates the update through a parameter-free, state-dependent projection. Since these modulation mechanisms are designed specifically to regulate repeated attention updates under looping, we apply them only within the looped stage \(\mathcal{B}\) of Looped-DiT.

\begin{table}[t]
\vspace{-10pt}
\centering
\caption{\textbf{Comparison with state-of-the-art text-to-image models across six benchmarks.}}
\label{tab:main-results}

\begingroup
\fontsize{9.5pt}{10pt}\selectfont
\setlength{\tabcolsep}{2pt}
\renewcommand{\arraystretch}{1.08}
\renewcommand{\tabularxcolumn}[1]{m{#1}}

\newcommand{\widehead}[1]{\makebox[0pt][c]{#1}}

\begin{tabularx}{\linewidth}{
    @{}
    >{\raggedright\arraybackslash}m{0.22\linewidth}
    *{8}{>{\centering\arraybackslash}X}
    @{}
}
\toprule

Model
& Params
& GenEval
& DPG
& PRISM
& CoRe
& \widehead{Spatial}
& \widehead{TIIF-Short}
& Average \\

\midrule

\multicolumn{9}{@{}l}
{\textcolor{gray}{\itshape Non-CoT models}} \\
\noalign{\vskip -6pt}
\multicolumn{9}{@{}c@{}}{%
    \textcolor{gray}{\hdashrule{\linewidth}{0.5pt}{1.5pt 1.5pt}}%
} \\

E-MMDiT~\citep{shen2025emmdit}
& 0.30B & 69.1 & 80.4 & 51.6 & 30.7 & 45.7 & 63.4 & 56.8 \\

SANA-0.6B~\citep{xie2024sana}
& 0.59B & 65.8 & 83.3 & 56.4 & 39.8 & 47.4 & 69.4 & 60.4 \\

DeCo-XXL/16~\citep{ma2025deco}
& 1.1B & 82.0 & 82.0 & 52.6 & 34.8 & 49.8 & 70.4 & 61.9 \\

URSA-0.6B~\citep{deng2025ursa}
& 0.86B & 64.1 & 85.6 & 59.6 & 43.4 & 52.0 & 67.6 & 62.1 \\

TiM-T2I~\citep{wang2025transition}
& 0.87B & 82.8 & 83.2 & 51.4 & 36.7 & 48.2 & 70.6 & 62.2 \\

DreamLite~\citep{feng2026dreamlite}
& 0.39B & 71.6 & 85.2 & 55.5 & 42.2 & 53.9 & 72.6 & 63.5 \\

CogView4~\citep{zheng2024cogview3}
& 6.4B & 73.0 & 85.1 & 60.9 & 45.0 & 53.3 & 66.9 & 64.0 \\

MiniT2I-B/16~\citep{minit2i2026}
& 0.26B & 87.5 & 84.1 & 55.6 & 44.0 & 52.0 & 75.0 & 66.4 \\

MiniT2I-L/16~\citep{minit2i2026}
& 0.91B & 88.3 & 84.8 & 58.9 & 44.4 & 52.2 & 75.0 & 67.3 \\

UniLiP-3B~\citep{tang2025unilip}
& 1.6B & \textbf{90.3} & 83.4 & 58.5 & 45.0
& 51.6 & 78.2 & 67.8 \\

InternVL-U~\citep{tian2026internvlu}
& 1.7B & 85.0 & 85.2 & 63.5 & 48.2
& 54.5 & 77.7 & 69.0 \\

\midrule

\multicolumn{9}{@{}l}
{\textcolor{gray}{\itshape CoT-Reasoning models}} \\
\noalign{\vskip -6pt}
\multicolumn{9}{@{}c@{}}{%
    \textcolor{gray}{\hdashrule{\linewidth}{0.5pt}{1.5pt 1.5pt}}%
} \\

GoT-R1~\citep{duan2025gotr1}
& 6.9B & 72.5 & 84.0 & 56.4 & 42.2 & 50.0 & 74.1 & 63.2 \\

T2I-R1~\citep{jiang2025t2ir1}
& 6.9B & 78.8 & 84.7 & 57.1 & 35.5 & 49.7 & 77.6 & 63.9 \\

Uni-CoT~\citep{qin2025unicot}
& 6.6B & 81.2 & 84.1 & 58.8 & 45.8 & 53.8 & 76.8 & 66.8 \\

\midrule

\cellcolor{cellblue}{\textbf{Looped-DiT B/16 (Ours)}}
& \cellcolor{cellblue}{\textbf{0.26B}}
& \cellcolor{cellblue}{87.4}
& \cellcolor{cellblue}{\textbf{87.0}}
& \cellcolor{cellblue}{\textbf{67.0}}
& \cellcolor{cellblue}{\textbf{53.5}}
& \cellcolor{cellblue}{\textbf{54.6}}
& \cellcolor{cellblue}{\textbf{79.7}}
& \cellcolor{cellblue}{\textbf{71.5}} \\

\bottomrule
\end{tabularx}
\endgroup
\end{table}

\section{Experiments}
\label{sec:exps}

In this section, we systematically evaluate Looped-DiT. We first examine its parameter efficiency, inference efficiency, and reasoning capability across varying loop depths. We then analyze whether alternative scaling strategies can reproduce the benefits of looping. Finally, we ablate Deep Supervision and Self-Modulating Attention. 

We base Looped-DiT on MiniT2I~\citep{minit2i2026}, a minimal pixel-space MMDiT, and train two variants, which we denote as B/16 and B/32 according to their patch sizes. We divide the 17 MMDiT blocks in the model into a split of $6, 5, 6$, and loop the middle 5 blocks for $N=4$ times. We evaluate our models on both general and reasoning-related datasets, including DPG-Bench~\citep{hu2024ella}, PRISM~\citep{fang2026flux}, T2I-CoReBench~\citep{li2026easier}, SpatialGenEval~\citep{wang2026everything}, GenEval~\citep{Ghosh2023GenEvalAO}, and TIIF-Short~\citep{wei2025tiif}. Unless otherwise specified, we use B/16 for the main results in Sec.~\ref{main-results} and the more lightweight B/32 for the analyses in Sec.~\ref{analysis} and ablations in Sec.~\ref{ablation}. When avg. results are reported, they are computed over all 6 datasets for B/16 and 4 reasoning-related datasets for B/32. Full implementation details for all these experiments are provided in App.~\ref{app:implementation}.

\subsection{Main Results}
\label{main-results}

\begin{figure}[t]
\vspace{-30pt}
\centering
\includegraphics[width=\linewidth]{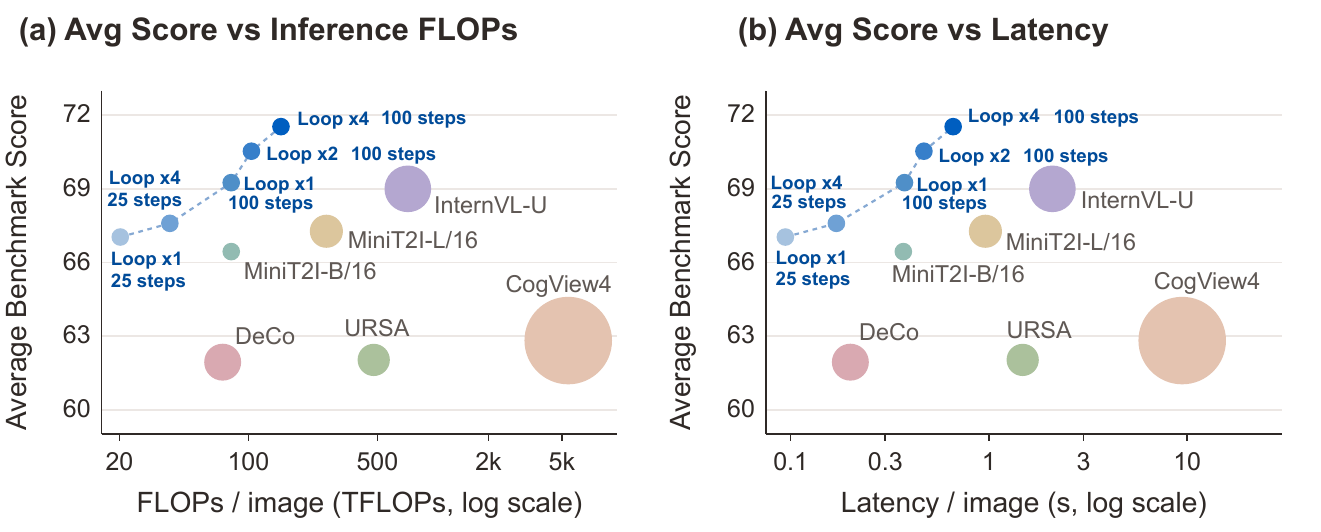}
\caption{\textbf{Performance-efficiency trade-offs across state-of-the-art text-to-image models.} Average Score denotes the mean performance across 6 benchmarks. Blue circles denote Looped-DiT B/16 with varying loop depth and denoising steps.}
\label{fig:main-flops-latency}
\end{figure}

\begin{figure}[!h]
\centering
\includegraphics[width=\linewidth]{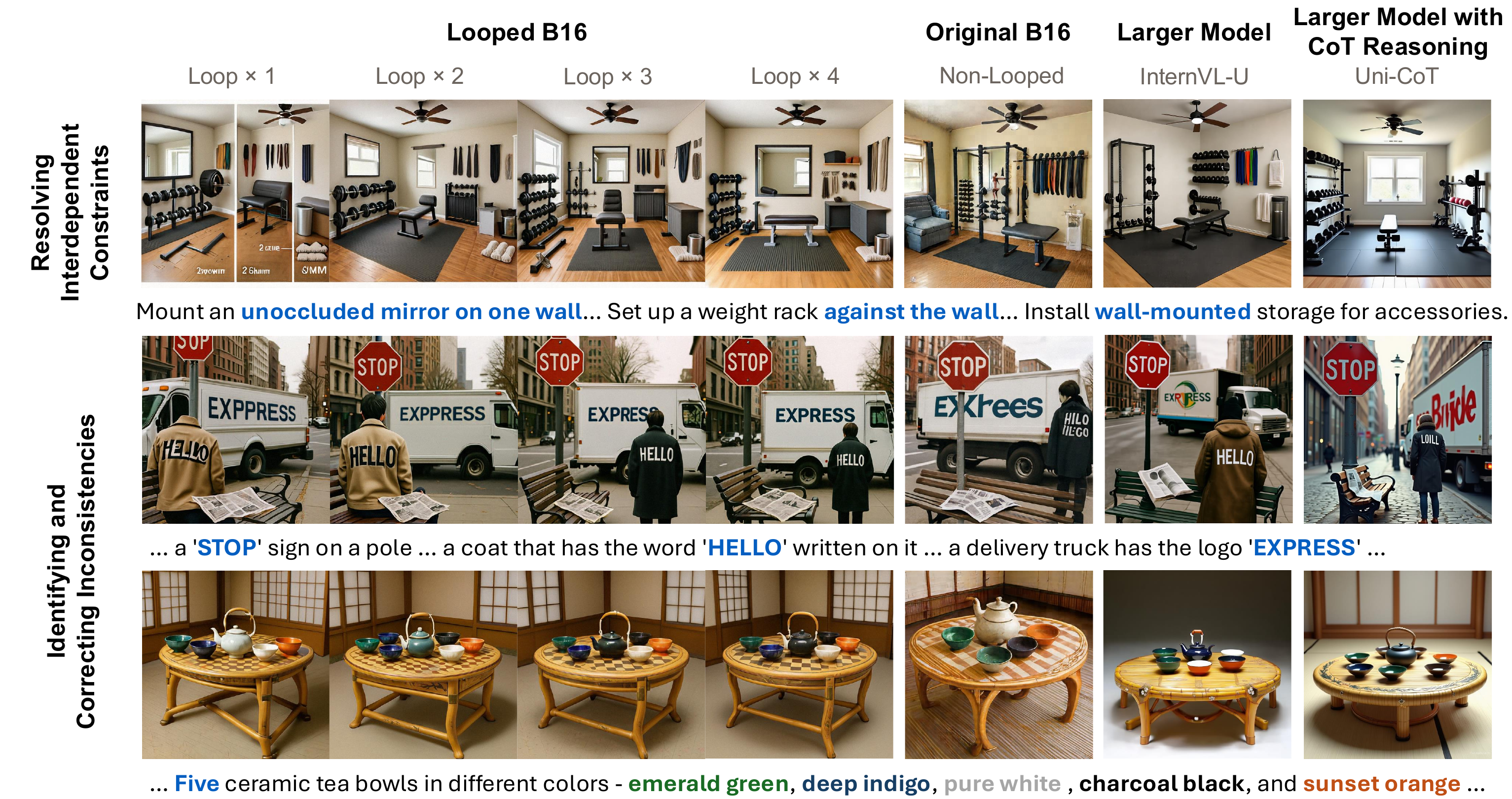}
\caption{\textbf{Qualitative results of Looped-DiT B/16.} Across successive loops, the model resolves spatial constraints and corrects inconsistencies.}
\label{fig:main-vis}
\end{figure}

\paragraph{Parameter efficiency.}
Tab.~\ref{tab:main-results} compares Looped-DiT B/16 with state-of-the-art text-to-image models under their official inference settings. As shown in the table, Looped-DiT B/16 achieves the best results on DPG-Bench, PRISM, T2I-CoReBench, SpatialGenEval and TIIF-Short, while using substantially fewer parameters than competing models. Notably, compared with the next-best model, InternVL-U, Looped-DiT B/16 achieves a 2.5-point higher avg. score with \(6.5\times\) fewer parameters.

\paragraph{Inference efficiency.}
Fig.~\ref{fig:main-flops-latency} compares the performance–efficiency trade-off between Looped-DiT B/16 and state-of-the-art text-to-image models in terms of inference compute and latency. Looped-DiT B/16 traces the Pareto frontier under both metrics, with loop depth and denoising steps providing flexible control over the inference budget.

\paragraph{Reasoning capability.}
Qualitatively, Looped-DiT B/16 exhibits behaviors consistent with constraint resolution and self-correction, two reasoning-related demands outlined in the introduction. As shown in Fig.~\ref{fig:main-vis}, successive loops introduce missing content, reorganize objects to satisfy spatial constraints, and correct rendering errors or remove extraneous objects, while such errors remain in the shown larger non-looped and explicit CoT baselines. Fig.~\ref{fig:main-flops-latency} further shows consistent performance gains as the loop count increases from 1 to 4, providing quantitative support for progressive refinement.

\subsection{Necessity of Looping}
\label{analysis}

Although looping improves performance and computational efficiency, it remains unclear whether these benefits are specific to looped computation or can be attained through alternative strategies. We therefore critically examine the necessity of looping here. Using the lightweight B/32 variant, we test whether its gains can be reproduced by increasing model depth or width, taking additional denoising steps, or introducing explicit chain-of-thought reasoning.

\definecolor{cellpeach}{RGB}{247,233,218}     
\definecolor{cellblue}{RGB}{230,237,247}      
\definecolor{cellmint}{RGB}{231,241,232}      

\begin{table}[h]
\vspace{15pt}
\centering

\caption{%
\textbf{Benefits of looping under matched parameter and different compute budgets.}
}
\label{tab:cmp-param-compute}

\begingroup
\setlength{\tabcolsep}{2pt}
\renewcommand{\arraystretch}{1.0}

\setlength{\aboverulesep}{2.5pt}
\setlength{\belowrulesep}{2.5pt}

\begin{tabular*}{\linewidth}{
    @{\extracolsep{\fill}}
    l c c c c c c c c
    @{}
}
\toprule

Model
& \makecell[c]{Train\\GFLOPs}
& \makecell[c]{Inference\\GFLOPs}
& \makecell[c]{Params\\(M)}
& \makecell[c]{Effective\\depth}
& \makecell[c]{Hidden\\dim}
& DeepSup
& Looping
& Avg. $\uparrow$ \\

\midrule

MiniT2I B/32
& 441
& 146
& \cellcolor{cellpeach}{260}
& 17
& 768
& \xmark
& \xmark
& 55.2 \\

\midrule

Deeper MiniT2I B/32
& 809
& \cellcolor{celllavender}{267}
& 473
& 32
& 768
& \xmark
& \xmark
& 57.7 (+2.6) \\

Wider MiniT2I B/32
& 805
& \cellcolor{celllavender}{268}
& 489
& 17
& 1056
& \xmark
& \xmark
& 58.6 (+3.5) \\

\midrule

Deeper MiniT2I B/32 w. DeepSup
& \cellcolor{cellmint}{1,246}
& \cellcolor{celllavender}{267}
& 473
& 32
& 768
& \cmark
& \xmark
& 58.1 (+3.0) \\

\midrule

Looped-DiT B/32 (Ours)
& \cellcolor{cellmint}{1,246}
& \cellcolor{celllavender}{267}
& \cellcolor{cellpeach}{260}
& 32
& 768
& \cmark
& \cmark
& \textbf{59.1 (+4.0)} \\

\bottomrule
\end{tabular*}
\vspace{5pt}

\endgroup
\end{table}

\paragraph{Do the gains from looping persist when controlling for model size and compute?}
Tab.~\ref{tab:cmp-param-compute} compares Looped-DiT B/32 with non-looped baselines under matched parameter, inference-compute, and training-compute settings. The deeper baseline replaces four passes through the five shared middle blocks with 20 distinct blocks to match the effective depth of 32, while the wider baseline increases the hidden dimension to approximately match the forward-pass compute. For the training-compute-matched baseline, we additionally apply Deep Supervision to the deeper model to match the training compute of Looped-DiT B/32. Looped-DiT B/32 improves the average score by 4.0 points over the parameter-matched baseline and outperforms both inference-compute-matched baselines despite using fewer parameters. Under matched training compute, it achieves 59.1 compared with 58.1 for the deeper baseline. These results show that the gains from looping persist across these parameter and compute settings.

\begin{figure}[h]
\vspace{15pt}
\centering
\includegraphics[width=\linewidth]{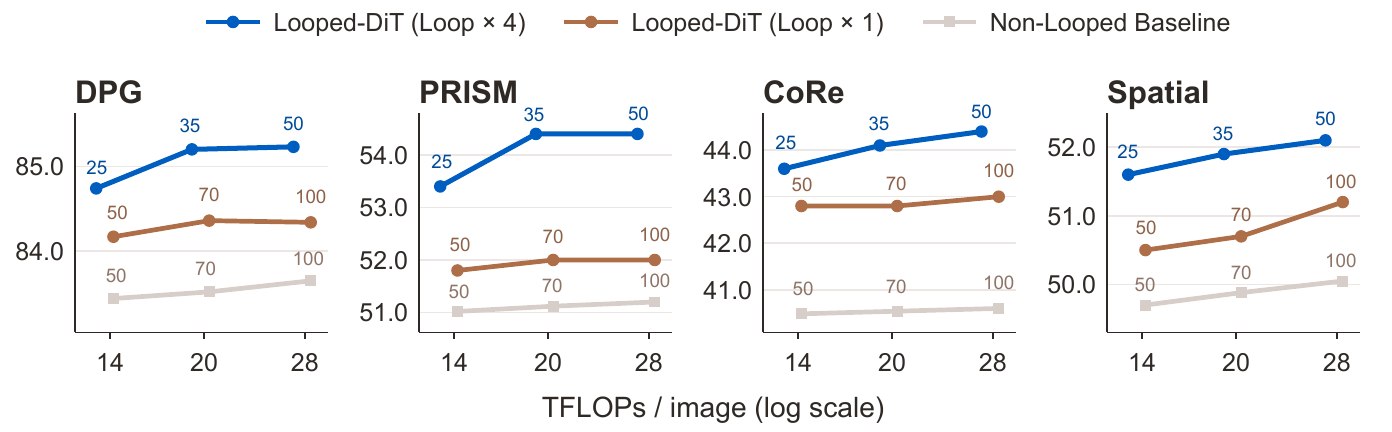}
\caption{%
\textbf{Loop iterations versus additional denoising steps.}
We compare 4-loop inference with single-pass inference using the same looped checkpoint or the non-looped model. Both single-pass settings use additional denoising steps to match the per-image inference FLOPs of 4-loop inference.%
}
\label{fig:cmp-step}
\end{figure}

\paragraph{Is increasing loop depth more effective than adding denoising steps?}
Fig.~\ref{fig:cmp-step} compares varying the loop depth of Looped-DiT B/32 with reallocating the same inference budget to additional denoising steps. We consider two non-looped baselines, one using the same checkpoint with looping disabled, thereby holding the learned weights and training history fixed, and the other using the same architecture trained without looping. From 25- to 50-step settings, allocating compute to loop depth consistently yields higher performance than allocating it to additional denoising steps, showing that looping is a more effective use of inference compute.

\begin{table}[t]
\centering

\newlength{\cottablewidth}
\newlength{\firstcolwidth}

\setlength{\cottablewidth}{0.60\linewidth}
\setlength{\firstcolwidth}{0.30\linewidth}

\newcolumntype{L}[1]{>{\raggedright\arraybackslash}p{#1}}
\newcolumntype{C}{>{\centering\arraybackslash}X}

\captionsetup{skip=\baselineskip}

\caption{\textbf{Complementary gains from looping and textual CoT}
over the non-looped model with original prompts.
Both combines looping with CoT rewriting.}
\label{tab:cmp-cot}

\setlength{\tabcolsep}{4pt}
\renewcommand{\arraystretch}{1.05}

\begin{tabularx}{\cottablewidth}{@{\hspace{10pt}}L{\firstcolwidth} C C C}
\toprule
Capability & CoT & Looping & Both \\
\midrule

Benchmark avg. & +2.1 & +4.0 & +5.5 \\

\midrule

\multicolumn{4}{@{}l@{}}{%
  \hspace{\tabcolsep}%
  \textcolor{gray}{%
    \itshape Constraint resolution (Looping-favored)%
  }%
} \\

\noalign{\vskip -7pt}

\multicolumn{4}{@{}l@{}}{%
  \hspace{\tabcolsep}%
  \textcolor{gray}{%
    \hdashrule{%
      \dimexpr\cottablewidth-2\tabcolsep\relax
    }{0.5pt}{1.5pt 1.5pt}%
  }%
  \hspace{\tabcolsep}%
} \\

\noalign{\vskip -3pt}

PRISM / Long Text     & +3.1 & +9.1 & +9.4 \\
CoRe / Multi-Relation & +1.2 & +8.3 & +8.4 \\
CoRe / Procedural     & +4.7 & +9.4 & +9.6 \\
Spatial / Orientation & +0.0 & +6.0 & +6.2 \\
Spatial / Motion      & +0.1 & +6.0 & +6.3 \\

\addlinespace[2.5pt]

\midrule

\multicolumn{4}{@{}l@{}}{%
  \hspace{\tabcolsep}%
  \textcolor{gray}{%
    \itshape Inferring implied visual content (CoT-favored)%
  }%
} \\

\noalign{\vskip -7pt}

\multicolumn{4}{@{}l@{}}{%
  \hspace{\tabcolsep}%
  \textcolor{gray}{%
    \hdashrule{%
      \dimexpr\cottablewidth-2\tabcolsep\relax
    }{0.5pt}{1.5pt 1.5pt}%
  }%
  \hspace{\tabcolsep}%
} \\

\noalign{\vskip -3pt}

CoRe / Generalization & +22.2 & +7.5 & +24.5 \\
CoRe / Hypothetical   & +9.4  & +3.5 & +11.6 \\
CoRe / Reconstructive & +6.2  & +3.4 & +8.9 \\

\bottomrule
\end{tabularx}

\end{table}

\paragraph{Is looping complementary to textual CoT?}
We compare CoT-based prompt rewriting for the non-looped model (CoT), looping with original prompts (Looping), and their combination (Both). For prompt rewriting, Qwen3~\citep{yang2025qwen3} is instructed to make the intended content explicit and clarify spatial layout, without adding unrelated details. Consistent with the constraint-resolution and self-correction behavior observed above, Tab.~\ref{tab:cmp-cot} shows larger gains from looping on subtasks that require joint satisfaction of relational, procedural, or spatial constraints. Adding CoT rewriting offers little further benefit on these subtasks.
Conversely, CoT yields larger gains on subtasks requiring the model to infer the implied visual content from examples and rules. Combining the two achieves the highest overall score, suggesting complementary strengths across different aspects of visual generation.

\vspace{0.5\baselineskip}
\subsection{Ablation Study}
\label{ablation}

\vspace{0.8\baselineskip}
\begin{table*}[h]
\centering
\captionsetup{skip=\baselineskip}
\caption{%
    \textbf{Component ablations.}
    Avg.\ is the mean over the four reported benchmarks.
    DeepSup denotes deep supervision; \xmark\ indicates final-loop only supervision.
    The loop weights $(w_1,w_2,w_3,w_4)$ are
    $(\sfrac{1}{8},\sfrac{1}{4},\sfrac{1}{2},1)$ for Exponential and
    $(\sfrac{1}{3},\sfrac{1}{3},\sfrac{1}{3},1)$ for Final + Mean.
    Gated and XSA are self-modulating attention variants.
}
\label{tab:abl-main}

\setlength{\tabcolsep}{6pt}
\renewcommand{\arraystretch}{1.0}

\begin{tabular*}{\textwidth}{
    @{\extracolsep{\fill}}
    c c c *{5}{c}
    @{}
}
\toprule

\multirow{2}{*}{Looping}
& \multirow{2}{*}{DeepSup}
& \multirow{2}{*}{Attention}
& \multicolumn{5}{c}{Benchmarks $\uparrow$} \\

\cmidrule(lr){4-8}

& & & DPG & PRISM & CoRe & Spatial & Avg. \\

\midrule

\xmark & \xmark & \xmark
& 82.0 & 51.2 & 39.2 & 48.3 & 55.2 \\

\cmark & \xmark & \xmark
& 84.4 & 51.1 & 39.3 & 49.9 & 56.2 \\

\cmark & Exponential & \xmark
& 84.3 & 52.1 & 41.0 & 51.0 & 57.1 \\

\cmark & Final + Mean & \xmark
& 84.4 & 53.9 & 40.6 & 50.9 & 57.5 \\

\cmark & \xmark & Gated
& 84.0 & 53.0 & 39.5 & 49.7 & 56.6 \\

\cmark & \xmark & XSA
& 84.2 & \textbf{54.5} & 43.2 & 52.1 & 58.5 \\

\cmark & Final + Mean & XSA
& \textbf{85.3} & 54.4 & \textbf{44.5}
& \textbf{52.3} & \textbf{59.1} \\

\bottomrule
\end{tabular*}
\vspace{0.6\baselineskip}

\end{table*}

\paragraph{Component ablations.}
Tab.~\ref{tab:abl-main} summarizes the contributions of looping, deep supervision, and self-modulating attention. Looping alone improves over the baseline model. For deep supervision, we compare two weighting schemes: Exponential assigns exponentially smaller weights to earlier loops, and Final + Mean adds the final-loop loss to the mean loss over earlier loops. Both outperform final-loop-only supervision, with Final + Mean performing better. For self-modulating attention, both Gated and XSA improve over unmodulated attention, with XSA providing larger gains. Combining Final + Mean supervision with XSA yields the highest average score.

\begin{figure}[t]
\centering
\includegraphics[width=\linewidth]{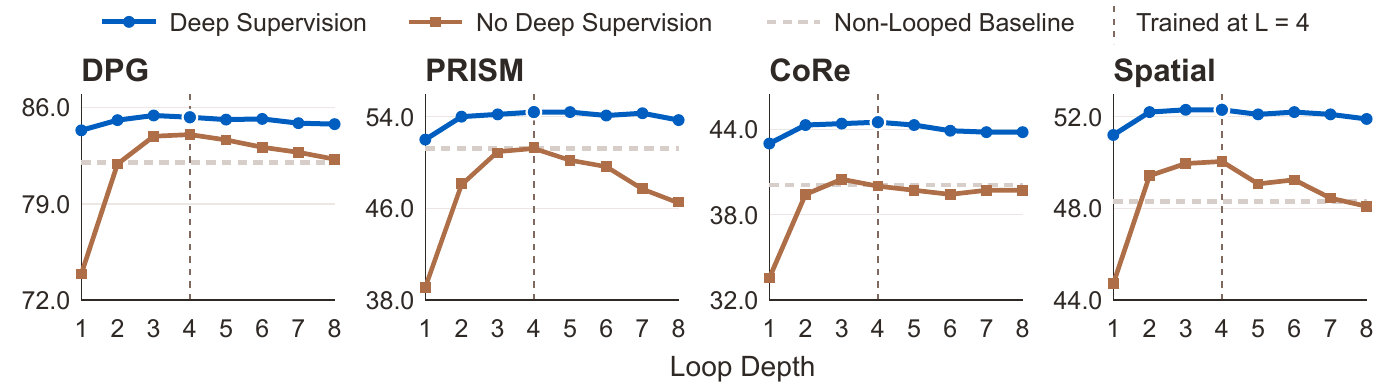}
\caption{\textbf{Deep supervision} improves the standard 4-loop result, stabilizes early exits, and maintains performance beyond the training loop count. The model trains at 4 loops and infers at 1--8 loops.
}
\label{fig:abl-deepsup}
\end{figure}

\begin{figure}[t]
\centering
\includegraphics[width=\linewidth]{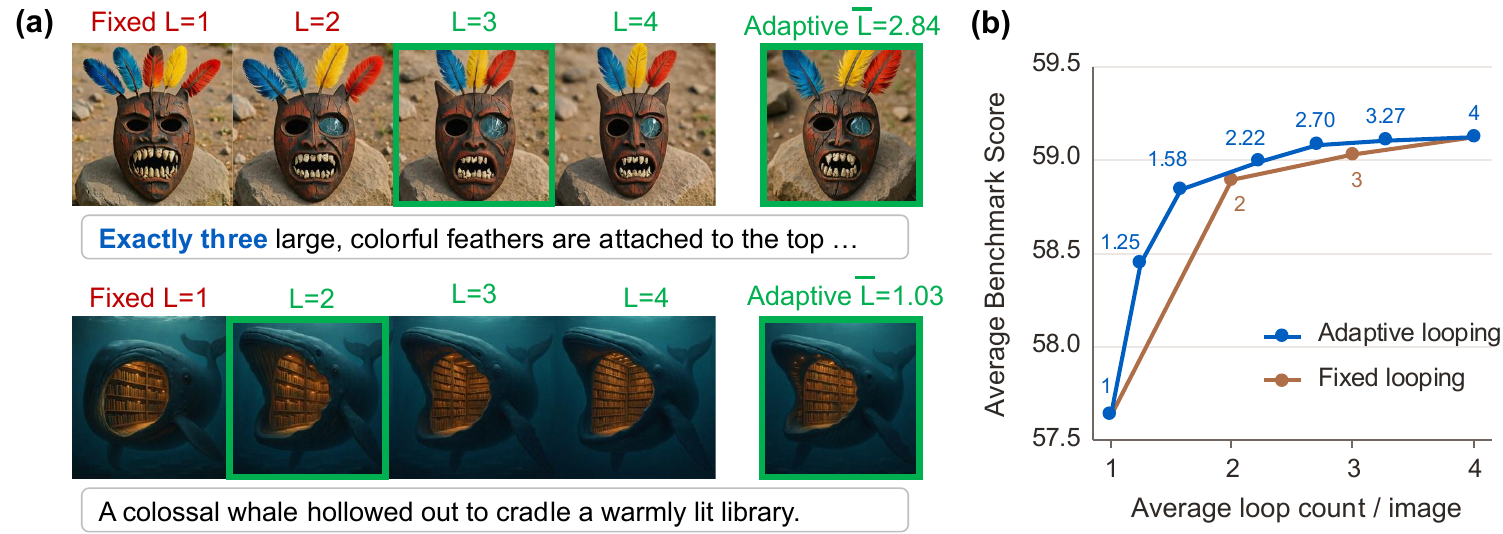}
\caption{\textbf{Adaptive looping results.} (a) By choosing different loop counts for each case, adaptive looping produces correct images with lower average loop loop counts compared to fixed looping. (b) Adaptive looping unlocks compute–performance trade-off without retraining, and also reaches better performance with fewer per-image loops. The numbers denote average per-image loops.
}
\label{fig:abl-adaptive}
\end{figure}

\paragraph{Deep supervision improves both single-pass and multi-loop performance.} Fig.~\ref{fig:abl-deepsup} shows that deep supervision improves performance across 1--8 inference loops, substantially reduces the penalty for early exit, and maintains performance beyond the 4 loops used during training. Notably, even the single-loop setting of the loop-trained model outperforms the standard non-looped baseline. Additional inference loops provide further gains with the same model.

The model's robust performance across different loop counts also enables \textit{adaptive looping}, where a lightweight gating network is introduced to decide whether to exit after each loop. The network consists of a cross-attention layer that extracts information from post-loop activations using 4 learnable queries, followed by an MLP head that predicts the expected benefit of continuing after the $r$-th loop, denoted as $\hat{y}_r$. We define its training target as the largest achievable future loss reduction per additional loop:
\begin{equation}
 y_r=\max\left\{0,\max_{k=r+1,\ldots,N}
             \frac{\ell_r-\ell_k}{k-r}\right\},
 \qquad r=1,2,\ldots,N-1,
 \label{eq:adaptive-target}
\end{equation}
where $\ell_k$ denotes the reconstruction error between the $k$-th loop prediction and the ground-truth image.
The target does not assume that the last pass is the most accurate exit; if no later exit improves upon the current prediction, it is set to zero.
At inference time, we set a threshold $\lambda$ and continue looping while $\hat{y}_r>\lambda$, while stopping at the first loop for which $\hat{y}_r\leq\lambda$. By varying $\lambda$, we can adjust the average number of loops according to the computation budget without retraining.
Fig.~\ref{fig:abl-adaptive}(a) shows qualitative examples in which adaptive looping uses enough loops to produce correct images while avoiding additional computation that offers little benefit once the prompt requirements are satisfied.
Fig.~\ref{fig:abl-adaptive}(b) shows quantitative results that adaptive looping outperforms fixed-loop inference at matched mean loop counts, halves the performance drop in a few-loop setting (at 1.25 loops), and reaches a comparable performance plateau earlier (at 2.70 loops).

\begin{wraptable}{r}{0.44\textwidth}
\vspace{0.5\baselineskip}
\centering

\caption{
    \textbf{Gains from XSA with and without looping.}
    Non-looped depths 17 and 32 match the looped model's
    parameter count and forward FLOPs, respectively.
    $\Delta_{\mathrm{XSA}}$ is the point gain over the paired
    variant without XSA.
}
\label{tab:abl-xsa}

\setlength{\tabcolsep}{2pt}
\renewcommand{\arraystretch}{1.0}

\begin{tabular*}{\linewidth}{
    @{\extracolsep{\fill}}
    c c c c c c
    @{}
}
\toprule

Looping
& \makecell{Effective\\depth}
& \makecell{Deep\\Sup}
& Attention
& \makecell{Avg.\\score $\uparrow$}
& $\Delta_{\mathrm{XSA}}$ \\

\midrule

\multicolumn{6}{@{}l@{}}{%
  \hspace{\tabcolsep}%
  \textcolor{gray}{\itshape Non-looped, parameter-matched}%
} \\
\noalign{\vskip -5pt}
\multicolumn{6}{@{}c@{}}{%
  \textcolor{gray}{\hdashrule{\linewidth}{0.5pt}{1.5pt 1.5pt}}%
} \\

\xmark & 17 & \xmark & \xmark
& 55.2 & \\

\xmark & 17 & \xmark & XSA
& 55.1 & $-0.1$ \\

\midrule

\multicolumn{6}{@{}l@{}}{%
  \hspace{\tabcolsep}%
  \textcolor{gray}{\itshape Non-looped, compute-matched}%
} \\
\noalign{\vskip -5pt}
\multicolumn{6}{@{}c@{}}{%
  \textcolor{gray}{\hdashrule{\linewidth}{0.5pt}{1.5pt 1.5pt}}%
} \\

\xmark & 32 & \cmark & \xmark
& 58.1 & \\

\xmark & 32 & \cmark & XSA
& 58.8 & $+0.7$ \\

\midrule

\multicolumn{6}{@{}l@{}}{%
  \hspace{\tabcolsep}%
  \textcolor{gray}{\itshape Looped}%
} \\
\noalign{\vskip -5pt}
\multicolumn{6}{@{}c@{}}{%
  \textcolor{gray}{\hdashrule{\linewidth}{0.5pt}{1.5pt 1.5pt}}%
} \\

\cmark & 32 & \cmark & \xmark
& 57.5 & \\

\cmark & 32 & \cmark & XSA
& \textbf{59.1} & $\mathbf{+1.6}$ \\

\bottomrule
\end{tabular*}

\end{wraptable}

\begin{figure}[t]
\centering
\vspace{-5pt}
\includegraphics[width=0.95\linewidth]{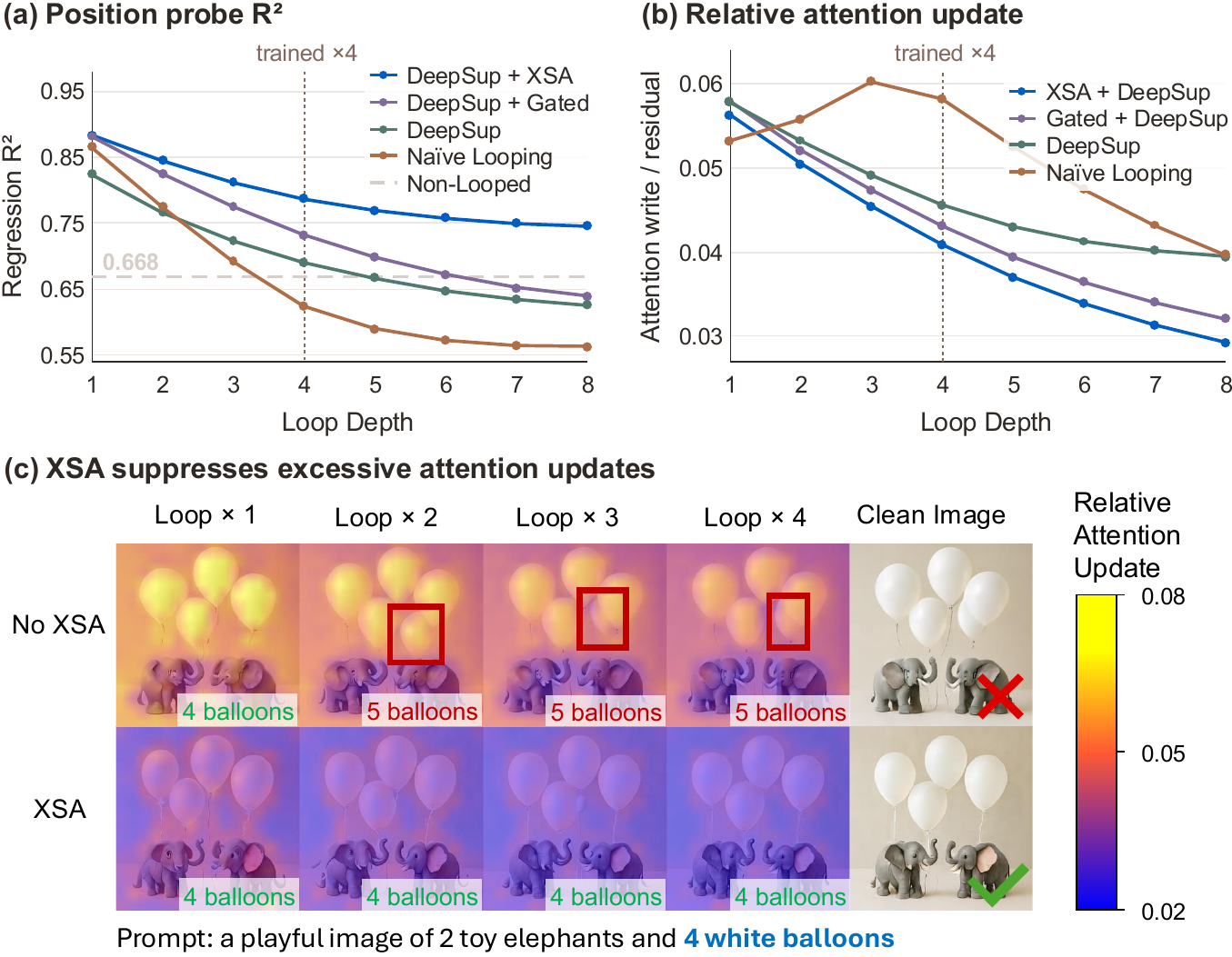}
\caption{
\textbf{Self-modulating attention reduces excessive writing.}
(a) Ridge-regression $R^2$ for decoding token positions from post-loop activations.
(b) Attention-update norm relative to the residual-stream norm. 
(c) An example of XSA reducing redundant attention updates, and thus preventing an error that would be introduced by looping without XSA at the same denoising step.
}
\label{fig:abl-attention}
\end{figure}

\paragraph{Self-modulating attention mitigates excessive updates across loops.}
Tab.~\ref{tab:abl-xsa} shows that XSA yields a larger gain for the looped model ($+1.6$) than for the parameter-matched ($-0.1$) and compute-matched ($+0.7$) non-looped baselines, suggesting that modulation is especially important under repeated updates. We hypothesize that unmodulated attention produces excessive updates across loops that progressively overwrite critical token information. Fig.~\ref{fig:abl-attention}(a) and Fig.~\ref{fig:abl-attention}(b) support this hypothesis. As loop count increases, unmodulated attention exhibits the largest relative update norm and the strongest decline in token-position decodability, measured by $R^2$, while XSA maintains the smallest updates and the highest position decodability, with Gated Attention in between. This ordering also matches their benchmark performance in Tab.~\ref{tab:abl-main}. Fig.~\ref{fig:abl-attention}(c) shows the same effect qualitatively. Without modulation, later loops continue modifying an already-correct image and introduce an extraneous object, whereas XSA suppresses later updates and preserves the existing structure. Together, these results suggest that self-modulating attention limits excessive updates across loops and helps preserve critical token information.

\section{Related Works}
\label{sec:related_works}

We summarize the most relevant work here and provide a more comprehensive discussion in Appendix~\ref{app:additional_related_work}. Looped Transformers~\citep{dehghani2018universal,geiping2026scaling} have been widely studied in language modeling as a way to increase computational depth by repeatedly applying shared blocks without adding parameters. Closest to our setting, Elastic Looped Transformers~\citep{goyal2026eltelasticloopedtransformers} apply this principle to class-conditional image and video generation, allowing loop depth to vary at inference. We build on MMDiT~\citep{esser2024scaling} to study loop-depth scaling in text-to-image generation, focusing on parameter efficiency and compositional and spatial reasoning. Beyond parameter efficiency, we also examine how looping changes the allocation of inference compute. This connects to few-step and one-step methods~\citep{luo2023latent,yin2024one}, which improve inference efficiency by reducing the number of denoising steps. Under a fixed inference budget, we find that increasing loop depth yields larger gains than adding more denoising steps, suggesting that looped computation can provide a more effective form of iterative computation for diffusion models. Together, these connections position our work at the intersection of Looped Transformers, text-to-image diffusion models, and efficient inference.
\section{Conclusions}
\label{sec:conlusions}

We introduce Looped-DiT, which repeatedly applies shared Transformer blocks within each denoising step to increase computational depth without increasing parameter count. To our knowledge, this is the first systematic investigation of looped computation for text-to-image generation. We find that naive looping is unreliable and progressively degrades spatial information, highlighting the need for intermediate supervision and adaptive regulation for effective looped computation. Importantly, controlled comparisons indicate that the resulting gains cannot be simply explained by scaling model depth or width, allocating the same inference compute to additional denoising steps, or introducing explicit textual reasoning. Beyond these performance gains, successive loops exhibit behaviors suggestive of latent visual reasoning, with iterative refinement of hidden representations progressively resolving constraints and correcting earlier errors. Together, these findings establish looping as a promising parameter-efficient way to improve future text-to-image models.

\clearpage

\bibliographystyle{plainnat}
\bibliography{main}

\clearpage


\beginappendix
\section{Appendix}

\subsection{Implementation Details}
\label{app:implementation}

\begin{table}[h]
\centering
\caption{\textbf{Model configurations.} Looped-DiT B/32 and Looped-DiT B/16 share the same Transformer backbone and differ mainly in patch size and the resulting image-token sequence length.}
\label{tab:model_config}
\small
\begin{tabular}{lcc}
\toprule
\textbf{Configuration} & \textbf{Looped-DiT B/32} & \textbf{Looped-DiT B/16} \\
\midrule
Image resolution & $512\times512$ & $512\times512$ \\
Patch size & 32 & 16 \\
Image tokens & 256 & 1024 \\
Joint sequence length & 512 & 1280 \\
Hidden dimension & 768 & 768 \\
Unique MMDiT blocks & 17 & 17 \\
Pre / loop / post blocks & $6/5/6$ & $6/5/6$ \\
Training loop depth & 4 & 4 \\
Effective block applications & 32 & 32 \\
Attention heads & 12 & 12 \\
Head dimension & 64 & 64 \\
SwiGLU hidden dimension & 2048 & 2048 \\
Text preamble blocks & 2 & 2 \\
Patch-embedding bottleneck & 128 & 128 \\
Maximum text length & 256 & 256 \\
Parameters & 260M & 260M \\
\bottomrule
\end{tabular}
\end{table}

\paragraph{Model architecture.}
We adopt MiniT2I~\citep{minit2i2026}, a pixel-space denoiser based on the MMDiT architecture~\citep{esser2024scaling}, as our backbone. Its simple architecture and training pipeline provide a controlled setting for studying loop depth. Image and text streams are processed through separate pathways and interact through joint attention. We retain the original embedding layers, pre-RMSNorm, query/key normalization, and rotary position embeddings, with no explicit timestep conditioning. 

We train two Looped-DiT variants at \(512\times512\) resolution, denoted B/32 and B/16 according to their patch sizes. B/32 produces 256 image tokens, while B/16 produces 1024. The resulting models contain 260.2M and 258.1M parameters, respectively. Both comprise 17 MMDiT blocks with a hidden dimension of 768, 12 attention heads of dimension 64, and a SwiGLU feed-forward hidden dimension of 2048.

To introduce looped computation, we divide the 17 MMDiT blocks as evenly as possible into pre-loop, looped, and post-loop groups, yielding a \([6,5,6]\) split. Given our resource constraints, we use this partition and a training loop depth of four as simple defaults rather than exhaustively optimized choices. Alternative partitions and training loop depths are left for future exploration. During training, the middle five blocks are repeated four times with shared parameters. This yields 32 effective block applications per denoising step while retaining only 17 unique parameterized blocks. Self-Modulating Attention is applied within the looped blocks to regulate the strength of attention updates across loops. At inference, loop depth can be varied without changing the model parameters. Unless otherwise stated, we use an inference loop depth of four. We use B/16 for the main results and B/32 for ablation studies. Tab.~\ref{tab:model_config} summarizes the full model configurations.

\paragraph{Training objective.}
We train all our models directly in pixel space using a flow-matching objective~\citep{lipman2022flow}. Given a clean image $x$ and noise $\epsilon\sim\mathcal{N}(0,4I)$, we construct the noisy image as
\begin{equation}
x_t = t x + (1-t)\epsilon,
\end{equation}
where $t$ is sampled from a logit-normal distribution with $\mu=-0.8$ and $\sigma=0.8$. The target and predicted velocity fields are then defined as
\begin{equation}
v = \frac{x-x_t}{\max(1-t,0.05)},
\qquad
\hat{v} = \frac{\hat{x}_0-x_t}{\max(1-t,0.05)},
\end{equation}
and we minimize the mean-squared error $\|\hat{v}-v\|_2^2$. During training, we use a noise scale of 2.0 and drop the text condition with probability 0.1 to enable classifier-free guidance~\citep{ho2022classifier}.

To directly supervise intermediate loop states, we apply Deep Supervision to predictions at loop depths $1$, $2$, and $3$. Each intermediate hidden state is passed through the six shared post-loop blocks, followed by the final normalization and prediction layers, and is optimized against the same flow-matching target as the final prediction at loop depth $4$. Intuitively, the final prediction should receive greater weight because it is produced after the full sequence of loop refinements, whereas earlier predictions correspond to intermediate states. We compare two weighting schemes for the four loop predictions, namely \textit{final+mean} weighting $(\sfrac{1}{3}, \sfrac{1}{3}, \sfrac{1}{3}, 1)$ and \textit{exponential} weighting $(\sfrac{1}{8}, \sfrac{1}{4}, \sfrac{1}{2}, 1)$. Based on results with the B/32 architecture, we find that \textit{final+mean} weighting achieves the best performance and therefore adopt it for B/16. These intermediate predictions are used exclusively during training and incur no additional parameters or inference-time computation.

\begin{table}[t]
    \centering
    \caption{\textbf{Training hyperparameters.} Hyperparameters are shared across model scales and training stages unless otherwise noted.}
    \label{tab:training_config}
    \small

    \begin{tabular}{lcccc}
        \toprule
        & \multicolumn{2}{c}{\textbf{Looped-DiT B/32}}
        & \multicolumn{2}{c}{\textbf{Looped-DiT B/16}} \\
        \cmidrule(lr){2-3}
        \cmidrule(lr){4-5}
        \textbf{Hyperparameter}
        & \textbf{Pretrain}
        & \textbf{Fine-tune}
        & \textbf{Pretrain}
        & \textbf{Fine-tune} \\
        \midrule
        Training steps
        & 250K & 40K & 500K & 80K \\
        Global batch size
        & \multicolumn{4}{c}{1024} \\
        Nodes $\times$ GPUs
        & $2 \times 8$ & $2 \times 8$
        & $4 \times 8$ & $4 \times 8$ \\
        Optimizer
        & \multicolumn{4}{c}{AdamW} \\
        Adam $(\beta_1, \beta_2)$
        & \multicolumn{4}{c}{$(0.9,\, 0.95)$} \\
        Weight decay
        & \multicolumn{4}{c}{0} \\
        Peak learning rate
        & \multicolumn{4}{c}{$4 \times 10^{-4}$} \\
        Initial learning rate
        & $1 \times 10^{-6}$ & -- 
        & $1 \times 10^{-6}$ & --  \\
        Warmup steps
        & 5K & --  & 5K & --  \\
        LR schedule
        & Warmup $\to$ Constant
        & Constant
        & Warmup $\to$ Constant
        & Constant \\
        Gradient clipping
        & \multicolumn{4}{c}{0.1} \\
        EMA decay
        & \multicolumn{4}{c}{0.99995} \\
        Condition dropout
        & \multicolumn{4}{c}{0.10} \\
        Noise scale
        & \multicolumn{4}{c}{2.0} \\
        $t$ distribution
        & \multicolumn{4}{c}{$\operatorname{LogitNormal}(-0.8,\, 0.8)$} \\
        Deep-supervision weighting
        & \multicolumn{4}{c}{\textit{final+mean} $(\sfrac{1}{3},\,\sfrac{1}{3},\,\sfrac{1}{3},\,1)$} \\
        \bottomrule
    \end{tabular}
\end{table}

\paragraph{Optimization.}
We optimize all our models using AdamW~\citep{loshchilov2017decoupled} with a global batch size of 1024 and a peak learning rate of $4\times10^{-4}$. During pretraining, the learning rate is linearly increased from $10^{-6}$ to $4\times10^{-4}$ over the first 5K steps and remains constant thereafter. Fine-tuning continues directly at this constant learning rate without additional warmup. B/32 is pretrained for 250K steps and fine-tuned for an additional 40K steps, while B/16 is pretrained for 500K steps and fine-tuned for an additional 80K steps. Tab.~\ref{tab:training_config} summarizes the full optimization configuration.

\paragraph{Training data.}


We use the same training-data setup as MiniT2I~\citep{minit2i2026}, pretraining Looped-DiT on CC12M~\citep{changpinyo2021conceptual} and fine-tuning it on a mixture of BLIP3o-60K~\citep{chen2025blip3}, DALL-E 3~\citep{opendatasets-dalle-3-dataset}, and ShareGPT-4o-Image~\citep{chen2025sharegpt}. Based on published papers and publicly released code and data, Looped-DiT and MiniT2I use the least training data among the models compared in Tab.~\ref{tab:main-results}. Among the remaining models that disclose their training-set size, all use at least twice as much data, while the others do not report their training-set size. We apply the same preprocessing to all training images, resizing each image so that its shorter side is 512 pixels, center-cropping it to $512\times512$, and normalizing it to $[-1,1]$. We use no random flipping or additional image augmentation.

\paragraph{Evaluation benchmarks.}
We evaluate Looped-DiT B/16 on six complementary text-to-image benchmarks covering compositional alignment, instruction following, and visual reasoning. GenEval~\citep{Ghosh2023GenEvalAO} measures object-centric compositional alignment, including counting, color, and spatial relations, while DPG-Bench~\citep{hu2024ella} evaluates adherence to dense prompts containing multiple objects, attributes, and relationships. TIIF-Bench~\citep{wei2025tiif} evaluates fine-grained instruction following across prompts of varying complexity. We use only its short-prompt split because our small-scale training setup primarily uses short captions with a maximum length of 256 tokens, making the long-prompt setting less representative of our training regime. T2I-CoReBench~\citep{li2026easier} targets complex composition and multi-step reasoning, PRISM-Bench~\citep{fang2026flux} evaluates prompt-image alignment and reasoning across diverse challenging generation tasks, and SpatialGenEval~\citep{wang2026everything} focuses specifically on spatial understanding and reasoning in information-dense scenes. For the B/32 analyses and ablations, we use DPG-Bench, T2I-CoReBench, PRISM-Bench, and SpatialGenEval. We focus on these four benchmarks because they are more directly related to the aspects of visual reasoning central to our study, particularly inferring implied visual content and resolving interdependent compositional and spatial constraints. Unless otherwise specified, when average results are reported, they are computed over all six benchmarks for B/16 and these four benchmarks for B/32.

\paragraph{Inference and evaluation.}
Unless otherwise stated, we evaluate checkpoints obtained using an exponential moving average (EMA) of the model weights. We use Euler sampling with 100 denoising steps, classifier-free guidance~\citep{ho2022classifier} with a scale of 6.0, and loop depth \(N=4\). Sampling is initialized from \(\mathcal{N}(0,4I)\). To reduce variance from stochastic image generation, we report the average score over three independent evaluation runs with different sampling seeds. To study inference-time scaling, we additionally vary loop depth from \(1\) to \(8\) and the number of denoising steps under matched inference-compute budgets. For latency comparisons, all models are evaluated on a single NVIDIA H100 GPU with batch size 1. We discard the first few generations as warm-up and report the mean latency over the next 100 generations.

\paragraph{Training and inference cost.}
Tab.~\ref{tab:compute_cost} reports the training and inference costs of the model configurations considered in our experiments using the Looped-DiT B/32 backbone. The compute-matched comparisons in the main paper are based on inference rather than training compute. Deep Supervision increases training compute because the intermediate states at loop depths $1$--$3$ are decoded through the shared six-block post-loop stage. Our full model therefore requires 1,246 GFLOPs per sample per training step, compared with 809 and 805 GFLOPs for the Deeper and Wider baselines, respectively, or approximately $1.54\times$ their training compute. This overhead is confined to training. At inference, the intermediate predictions are not evaluated, so Deep Supervision adds neither parameters nor inference compute. The full model thus retains the inference cost of the corresponding looped model and approximately matches the Deeper and Wider baselines. Moreover, this additional cost is incurred only during training and is amortized over subsequent generations. XSA introduces negligible computational overhead and does not change the reported training or inference GFLOPs. Overall, our method concentrates its additional computation during training while preserving the inference efficiency of the looped model.

\begin{table}[t]
    \centering
    \caption{\textbf{Training and inference cost.}
    Training GFLOPs are measured per sample for one full training step, including forward and backward computation and, when applicable, the Deep Supervision exits. Inference GFLOPs are measured per denoising forward pass. Parenthesized values are relative to the parameter-matched MiniT2I-B/32 baseline.}
    \label{tab:compute_cost}
    \small
    \setlength{\tabcolsep}{3pt}

    \begin{tabular}{@{}lccc@{}}
        \toprule
        & & \textbf{Training} & \textbf{Inference} \\
        \cmidrule(lr){3-3}
        \cmidrule(lr){4-4}
        \textbf{Configuration}
        & \textbf{Params}
        & \textbf{GFLOPs}
        & \textbf{GFLOPs} \\
        \midrule

        Compute-matched (Deeper)
        & \cellcolor{costblue}473M $(1.82\times)$
        & \cellcolor{costblue}809 $(1.83\times)$
        & \cellcolor{costblue}267 $(1.83\times)$ \\

        Compute-matched (Wider)
        & \cellcolor{costblue}489M $(1.88\times)$
        & \cellcolor{costblue}805 $(1.83\times)$
        & \cellcolor{costblue}268 $(1.84\times)$ \\

        \midrule

        Parameter-matched (MiniT2I-B/32)
        & \cellcolor{costgreen}260M $(1.00\times)$
        & \cellcolor{costgreen}441 $(1.00\times)$
        & \cellcolor{costgreen}146 $(1.00\times)$ \\

        \quad + Looping
        & \cellcolor{costgreen}260M $(1.00\times)$
        & \cellcolor{costblue}809 $(1.83\times)$
        & \cellcolor{costblue}267 $(1.83\times)$ \\

        \quad + XSA
        & \cellcolor{costgreen}260M $(1.00\times)$
        & \cellcolor{costblue}809 $(1.83\times)$
        & \cellcolor{costblue}267 $(1.83\times)$ \\

        \quad + DeepSup
        & \cellcolor{costgreen}260M $(1.00\times)$
        & \cellcolor{costrose}1{,}246 $(2.83\times)$
        & \cellcolor{costblue}267 $(1.83\times)$ \\

        \quad + DeepSup + XSA (ours)
        & \cellcolor{costgreen}260M $(1.00\times)$
        & \cellcolor{costrose}1{,}246 $(2.83\times)$
        & \cellcolor{costblue}267 $(1.83\times)$ \\

        \bottomrule
    \end{tabular}
\end{table}

\subsection{Additional Related Work}
\label{app:additional_related_work}

\noindent \paragraph{Diffusion Transformers for Text-to-Image Generation.} DiT~\citep{peebles2023scalable} establishes Transformers~\citep{vaswani2017attention} as scalable backbones for diffusion models, motivating their adoption in text-to-image generation. PixArt-$\alpha$~\citep{chen2024pixart} demonstrates that Transformer-based text-to-image models can be trained efficiently at scale, while Stable Diffusion 3~\citep{esser2024scaling} introduces MMDiT with modality-specific parameters and joint attention over image and text tokens. SANA~\citep{xie2024sana} improves high-resolution generation efficiency through linear attention and highly compressed latent representations, whereas Qwen-Image~\citep{wu2025qwen} scales multimodal diffusion Transformers toward stronger text rendering and complex visual generation. In contrast to these efforts on architectural design, representation learning, and model scaling, we study repeatedly applying shared Transformer blocks as a parameter-efficient way to increase computational depth in text-to-image generative models.

\noindent \paragraph{Few-Step and One-Step Text-to-Image Generation.} The iterative sampling process of diffusion and flow models has motivated substantial work on reducing the number of model evaluations required for generation. Progressive distillation~\citep{salimans2022progressive} progressively compresses multi-step diffusion samplers, while Consistency Models~\citep{song2023consistency} learn mappings that support one- or few-step generation. Building on these approaches, text-to-image methods including Latent Consistency Models~\citep{luo2023latent}, InstaFlow~\citep{liu2023instaflow}, SDXL-Turbo~\citep{sauer2024adversarial}, SDXL-Lightning~\citep{lin2024sdxl}, DMD~\citep{yin2024one}, and Hyper-SD~\citep{ren2024hyper} further accelerate generation through consistency learning, flow-based formulations, adversarial distillation, or distribution matching. Recent work also develops objectives designed directly for one-step inference. MeanFlow~\citep{geng2026mean} learns average velocity over finite time intervals without pretrained teachers or distillation, while Drifting Models~\citep{deng2026generative} shift distribution evolution from iterative inference to training. In contrast to approaches that reduce the number of external refinement steps, we investigate whether iterative computation can instead be internalized through loop depth, providing a complementary axis for allocating inference compute between sampling steps and hidden-state refinement.

\noindent \paragraph{Looped Transformers.} Looped Transformers increase computational depth by repeatedly applying shared parameters. Early approaches such as Universal Transformers~\citep{dehghani2018universal} and ALBERT~\citep{lan2019albert} explored looped computation and cross-layer parameter sharing, while later work showed that looping can support iterative algorithm execution and in-context learning~\citep{giannou2023looped,yang2024looped}. More recent studies connect loop depth to reasoning, showing that additional hidden-state computation can complement or substitute for explicit chain-of-thought generation~\citep{saunshi2025reasoning,xu2025formal,geiping2026scaling}. Related work on latent and looped computation includes Coconut~\citep{hao2024training}, which performs reasoning in continuous latent states, Relaxed Recursive Transformers~\citep{bae2025relaxed}, which introduce greater flexibility in parameter-shared depth, and Mixture-of-Recursions~\citep{bae2026mixture}, which adaptively allocates recursion depth across tokens. Other work extends looped computation with explicit state memory in MeSH~\citep{yu2026mesh}, multi-resolution processing in SpiralFormer~\citep{yu2026spiralformer}, and recurrence within mixture-of-experts models in SMELT~\citep{wang2026smelt}. Closest to our setting, Elastic Looped Transformers (ELT)~\citep{goyal2026eltelasticloopedtransformers} study looped computation for class-conditional image and video generation, with an emphasis on varying the number of loop iterations at inference time. They observe a similar degradation at early loop exits and address it through intra-loop self-distillation, where shallower loop configurations are trained to match the maximum-loop configuration. In contrast, our Deep Supervision directly optimizes intermediate-loop predictions with the same training objective as the final prediction, without requiring a teacher configuration or distillation objective. Our setting also differs fundamentally from ELT, which does not consider text-to-image generation. We extend looped computation to open-ended text-to-image generation, where the model must interpret and satisfy diverse natural-language constraints rather than a single class label, and investigate whether repeated hidden-state refinement can support latent visual reasoning. To our knowledge, this is the first systematic study of looped computation for text-to-image generation.

\subsection{Additional Details on Self-Modulating Attention}
\label{app:sma}

We provide additional details for the two realizations of Self-Modulating Attention~(SMA) used in Looped-DiT: Gated Attention~\citep{qiu2026gated} and Exclusive Self Attention (XSA)~\citep{zhai2026exclusive}.

\paragraph{Placement within the attention block.}
SMA is applied after scaled dot-product attention and before head concatenation and output projection. For token $i$ and attention head $h$, standard attention produces
\begin{equation}
    o_{i,h}
    =
    \sum_j \alpha_{ij,h} v_{j,h},
\end{equation}
where the sum is taken over the joint sequence of image and text tokens. SMA transforms $o_{i,h}$ into a modulated output $z_{i,h}$. The outputs from all heads are then concatenated and passed through the standard output projection,
\begin{equation}
    \Delta_i
    =
    W_O
    \operatorname{Concat}
    \left(
        z_{i,1}, \ldots, z_{i,H}
    \right).
\end{equation}
Thus, SMA changes what each attention head writes to the residual stream without modifying the attention weights themselves.

We apply SMA only within the looped stage $\mathcal{B}$, while the pre-loop and post-loop stages use standard attention. Since the parameters of $\mathcal{B}$ are shared across loop iterations, any learned SMA parameters are shared as well. The modulation itself nevertheless changes across loops because it is recomputed from the current hidden states.

\paragraph{Gated Attention.}
For Gated Attention, let $u_i$ denote the normalized hidden state of token $i$. The modulation for head $h$ is a token-dependent scalar gate,
\begin{equation}
    G_{i,h}^{\mathrm{gate}}
    =
    \sigma\left(
        w_{g,h}^{\top}u_i + b_{g,h}
    \right),
\end{equation}
where $w_{g,h}$ and $b_{g,h}$ are modality-specific gate parameters, with separate parameters for image and text tokens. The output of head $h$ is then modulated as
\begin{equation}
    z_{i,h}^{\mathrm{gate}}
    =
    G_{i,h}^{\mathrm{gate}} o_{i,h}.
\end{equation}
The gate therefore explicitly controls the magnitude of each head contribution before head concatenation and output projection.

\paragraph{Exclusive Self Attention.}
XSA provides a parameter-free realization of SMA. For token $i$ and head $h$, let
\begin{equation}
    \hat v_{i,h}
    =
    \frac{v_{i,h}}
    {\lVert v_{i,h}\rVert_2}
\end{equation}
denote the normalized value vector of the token itself. XSA defines the modulation
\begin{equation}
    G_{i,h}^{\mathrm{xsa}}
    =
    I-\hat v_{i,h}\hat v_{i,h}^{\top},
\end{equation}
which projects the attention output onto the subspace orthogonal to the token's own value direction. The modulated head output is
\begin{equation}
    z_{i,h}^{\mathrm{xsa}}
    =
    G_{i,h}^{\mathrm{xsa}} o_{i,h}.
\end{equation}
Since
\begin{equation}
    G_{i,h}^{\mathrm{xsa}} v_{i,h}=0,
\end{equation}
the direct self-value contribution is eliminated. Expanding the attention output gives
\begin{equation}
\begin{aligned}
    z_{i,h}^{\mathrm{xsa}}
    &=
    G_{i,h}^{\mathrm{xsa}}
    \sum_j \alpha_{ij,h}v_{j,h} \\
    &=
    \sum_{j\neq i}
    \alpha_{ij,h}
    G_{i,h}^{\mathrm{xsa}}v_{j,h}.
\end{aligned}
\end{equation}
This also shows that XSA is not equivalent to simply scaling the attention output by $1-\alpha_{ii,h}$. The projection removes not only the token's direct self-value contribution but also any component of the remaining value vectors that lies along the token's own value direction.

Because $G_{i,h}^{\mathrm{xsa}}$ is an orthogonal projection,
\begin{equation}
    \left\|
    z_{i,h}^{\mathrm{xsa}}
    \right\|_2
    \leq
    \left\|
    o_{i,h}
    \right\|_2,
\end{equation}
so XSA is non-expansive at the individual-head output before the output projection.

\paragraph{Comparison of modulation mechanisms.}
Gated Attention and XSA realize Self-Modulating Attention in different ways. Gated Attention explicitly controls the magnitude of each head output through a learned scalar gate, whereas XSA constrains the update through a parameter-free, state-dependent projection. Both mechanisms operate on the attention output before it is written back to the residual stream, and both are recomputed from the current representation at every loop iteration. This allows the shared looped blocks to adapt their attention updates as the hidden states evolve across repeated passes.

\subsection{Limitations and Future Work}

Our study has several limitations that suggest directions for future work. First, our
experiments focus on MiniT2I-based MMDiT models at approximately 260M parameters
and $512\times512$ resolution. While the results consistently support looped computation
in this controlled setting, it remains unclear how these findings scale to substantially
larger models, latent-space architectures, and higher-resolution generation. Evaluating
looped computation across these settings is an important direction for future work.

Second, due to resource constraints, we use a fixed $[6,5,6]$ pre-loop, looped, and
post-loop partition and a training loop depth of four rather than exhaustively exploring
the design space. Future work could study how loop placement, training depth, and the
fraction of shared blocks interact with model scale and inference budget.

Third, Deep Supervision increases training computation because intermediate loop states
are additionally decoded during optimization. Although this overhead is absent at
inference, reducing the training cost of intermediate supervision or developing more
efficient objectives for learning useful intermediate states would further improve the
overall efficiency of looped models.

Finally, our evidence for latent visual reasoning is primarily behavioral and
representational. The progressive correction of errors across loops is consistent with
iterative reasoning, but does not provide a complete mechanistic account of how these
behaviors emerge. More direct analyses of information flow and computation across loop
iterations may help clarify when iterative hidden-state refinement constitutes reasoning
and how such behavior changes with scale.

\end{document}